\documentclass[letterpaper, 10 pt, conference]{ieeeconf}
\IEEEoverridecommandlockouts                              
\usepackage{comment}
\usepackage{url}

\usepackage{multirow}
\usepackage{graphicx}
\usepackage{epsfig}
\usepackage{epic,eepic}
\usepackage{times}
\usepackage{mathptmx}
\usepackage{cite}
\usepackage{color}

\usepackage{times}

\definecolor{red}{rgb}{1,0,0}
\definecolor{green}{rgb}{0,1,0}
\definecolor{blue}{rgb}{0,0,1}
\definecolor{violet}{rgb}{1,0,1}
\definecolor{cyan}{cmyk}{1,0,0,0}
\definecolor{magenta}{cmyk}{0,1,0,0}
\definecolor{yellow}{cmyk}{0,0,1,0}

\definecolor{white}{rgb}{1,1,1}

\usepackage{arydshln}

\newcommand{\CommentOut}[1]{}

\newcommand{\noeditage}[1]{#1} \newcommand{\editage}[1]{}

\newcommand{\myauthor}[1]{\author{#1}}

\begin{document}

\newcommand{\FIG}[3]{
\begin{minipage}[b]{#1cm}
\begin{center}
\includegraphics[width=#1cm]{#2}\\
{\scriptsize #3}
\end{center}
\end{minipage}
}

\newcommand{\FIGU}[3]{
\begin{minipage}[b]{#1cm}
\begin{center}
\includegraphics[width=#1cm,angle=180]{#2}\\
{\scriptsize #3}
\end{center}
\end{minipage}
}

\newcommand{\FIGm}[3]{
\begin{minipage}[b]{#1cm}
\begin{center}
\includegraphics[width=#1cm]{#2}\\
{\scriptsize #3}
\end{center}
\end{minipage}
}

\newcommand{\FIGR}[3]{
\begin{minipage}[b]{#1cm}
\begin{center}
\includegraphics[angle=-90,width=#1cm]{#2}
\\
{\scriptsize #3}
\vspace*{1mm}
\end{center}
\end{minipage}
}

\newcommand{\FIGRpng}[5]{
\begin{minipage}[b]{#1cm}
\begin{center}
\includegraphics[bb=0 0 #4 #5, angle=-90,clip,width=#1cm]{#2}\vspace*{1mm}
\\
{\scriptsize #3}
\vspace*{1mm}
\end{center}
\end{minipage}
}

\newcommand{\FIGCpng}[5]{
\begin{minipage}[b]{#1cm}
\begin{center}
\includegraphics[bb=0 0 #4 #5, angle=90,clip,width=#1cm]{#2}\vspace*{1mm}
\\
{\scriptsize #3}
\vspace*{1mm}
\end{center}
\end{minipage}
}

\newcommand{\FIGpng}[5]{
\begin{minipage}[b]{#1cm}
\begin{center}
\includegraphics[bb=0 0 #4 #5, clip, width=#1cm]{#2}\vspace*{-1mm}\\
{\scriptsize #3}
\vspace*{1mm}
\end{center}
\end{minipage}
}

\newcommand{\FIGtpng}[5]{
\begin{minipage}[t]{#1cm}
\begin{center}
\includegraphics[bb=0 0 #4 #5, clip,width=#1cm]{#2}\vspace*{1mm}
\\
{\scriptsize #3}
\vspace*{1mm}
\end{center}
\end{minipage}
}

\newcommand{\FIGRt}[3]{
\begin{minipage}[t]{#1cm}
\begin{center}
\includegraphics[angle=-90,clip,width=#1cm]{#2}\vspace*{1mm}
\\
{\scriptsize #3}
\vspace*{1mm}
\end{center}
\end{minipage}
}

\newcommand{\FIGRm}[3]{
\begin{minipage}[b]{#1cm}
\begin{center}
\includegraphics[angle=-90,clip,width=#1cm]{#2}\vspace*{0mm}
\\
{\scriptsize #3}
\vspace*{1mm}
\end{center}
\end{minipage}
}

\newcommand{\FIGC}[5]{
\begin{minipage}[b]{#1cm}
\begin{center}
\includegraphics[width=#2cm,height=#3cm]{#4}~$\Longrightarrow$\vspace*{0mm}
\\
{\scriptsize #5}
\vspace*{8mm}
\end{center}
\end{minipage}
}

\newcommand{\FIGf}[3]{
\begin{minipage}[b]{#1cm}
\begin{center}
\fbox{\includegraphics[width=#1cm]{#2}}\vspace*{0.5mm}\\
{\scriptsize #3}
\end{center}
\end{minipage}
}

\title{\bf\Large%
Cross-view Self-localization from Synthesized Scene-graphs 
}

\myauthor{Ryogo Yamamoto ~~~~~~ Kanji Tanaka ~~~~
\thanks{Our work has been supported in part by JSPS KAKENHI Grant-in-Aid for Scientific Research (C) 20K12008 and 23K11270.}
\thanks{$*$%
R. Yamamoto, and K. Tanaka are with Robotics Coarse, 
Department of Engineering, University of Fukui, Japan. 
{\tt\small{\{mf220362@g. tnkknj@\}u-fukui.ac.jp}}
}}

\newcommand{\mybox}[3]{
\fbox{
\begin{tabular}{ll}
\begin{minipage}[b]{0.1cm}
~
\vspace*{#1cm}
\end{minipage}
&
\begin{minipage}[b]{#2cm}
\small \bf
#3
\end{minipage}
\end{tabular}
}
}

\newcommand{\figA}{
\begin{figure}[t]
\hspace*{2.0mm}
\FIG{3.8}{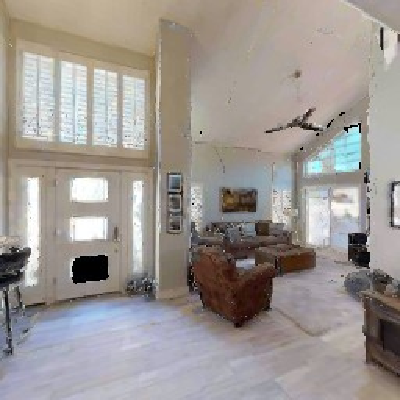}{}
\FIG{3.8}{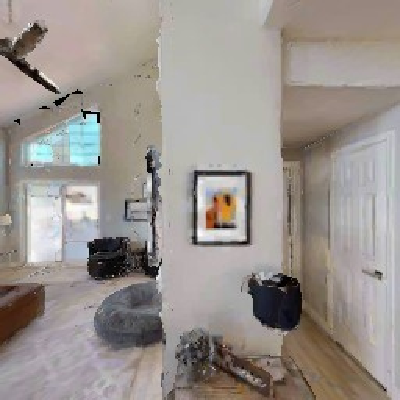}{}

\caption{%
Given only a single training image per place class, our goal is to achieve robust place classification at very different viewpoints. Shown in the figure is a typical example pairing of training and test images that belong to the same place class, in our challenging cross-view localization.
}\label{fig:A}
\end{figure}
}

\newcommand{\figB}{
\begin{figure}[t]

\FIG{8}{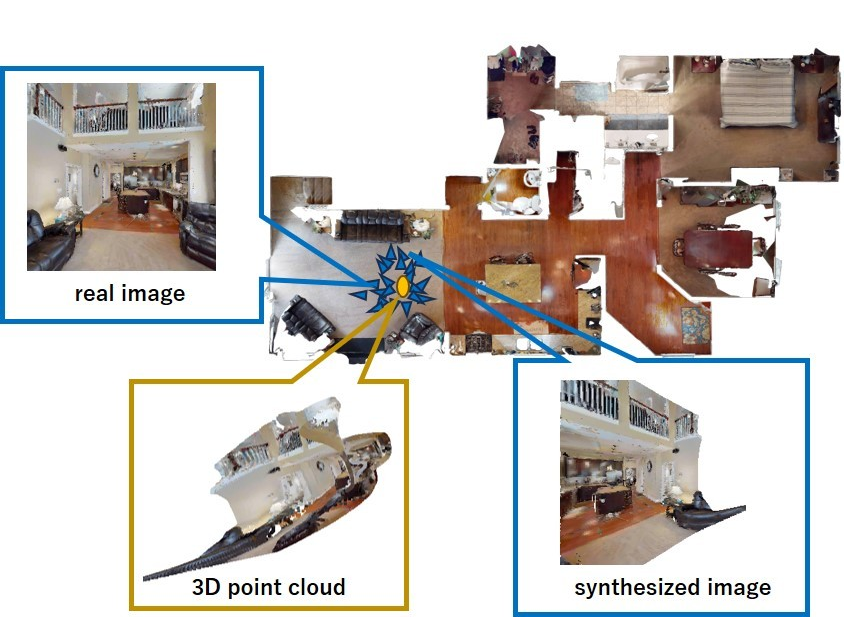}{}

\caption{%
System overview. For each place class, a number of virtual viewpoints are placed surrounding the place's representative point (orange circle), and a scene graph is synthesized for each virtual viewpoint from the only available real training image. The blue triangles are real and virtual viewpoints, and typical examples of real and synthesized images are visualized. The 3D point cloud is generated as an intermediate to synthesize the virtual viewpoint’s scene graph.
}\label{fig:B}
\end{figure}
}

\newcommand{\figC}{
\begin{figure}[t]
\hspace*{3mm}
\FIG{3.5}{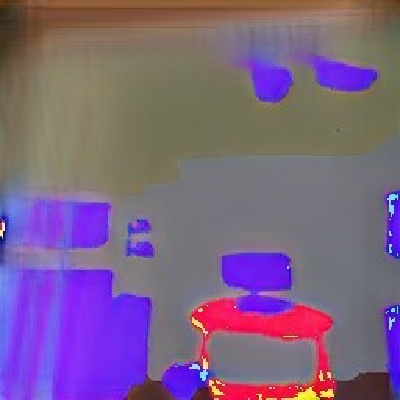}{}
\FIG{3.5}{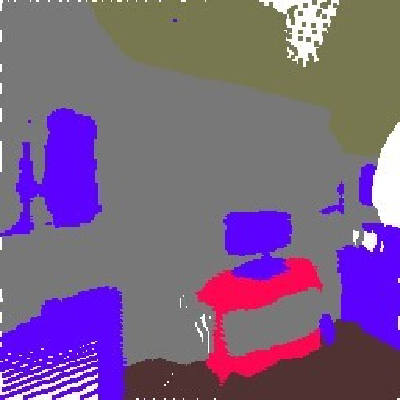}{}
\caption{%
View synthesis results by SynSin \cite{I14} (left) and the proposed view-synthesis (right).
}\label{fig:C}
\end{figure}
}

\newcommand{\figE}{
\begin{figure}[t]
  \begin{minipage}[b]{3cm}
   \FIG{3}{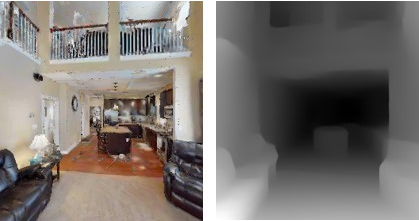}{(a)}\\
   \FIG{3}{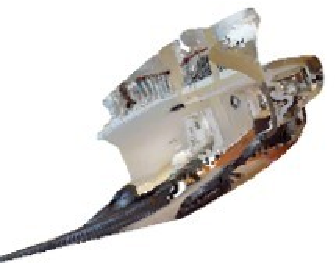}{(b)}
  \end{minipage}
  \begin{minipage}[b]{5cm}
   \FIG{5}{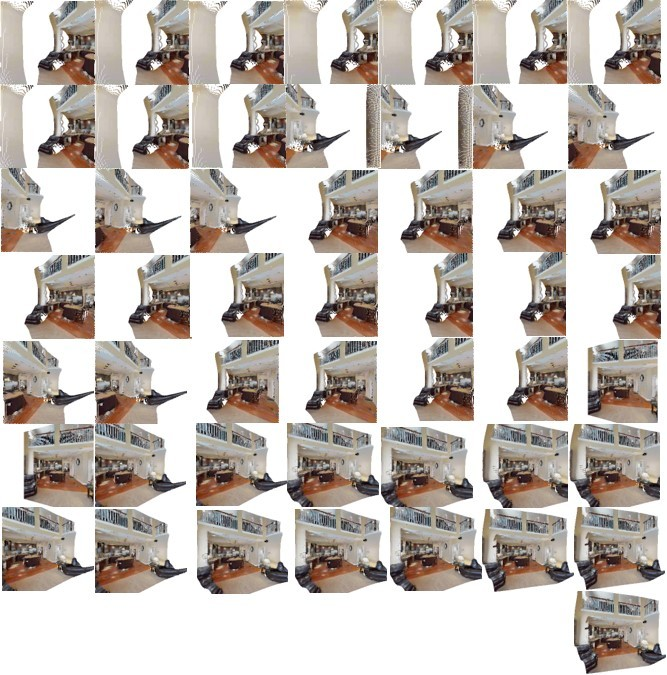}{(c)}
  \end{minipage}
\vspace*{-3mm}\\
\leftline{\scriptsize 00800-TEEsavR23oF}
\caption{%
View synthesis examples:
(a)
Input RGB image and monocular depth image calculated from it;
(b)
A 3D point cloud calculated from a monocular depth image viewed from another viewpoint;
(c)
Synthesized images corresponding to training synthesized scene graphs, for visualization purpose only.
}\label{fig:E}
\end{figure}
}

\newcommand{\figD}{
\begin{figure}[t]
\hspace*{3mm}
\FIG{2.3}{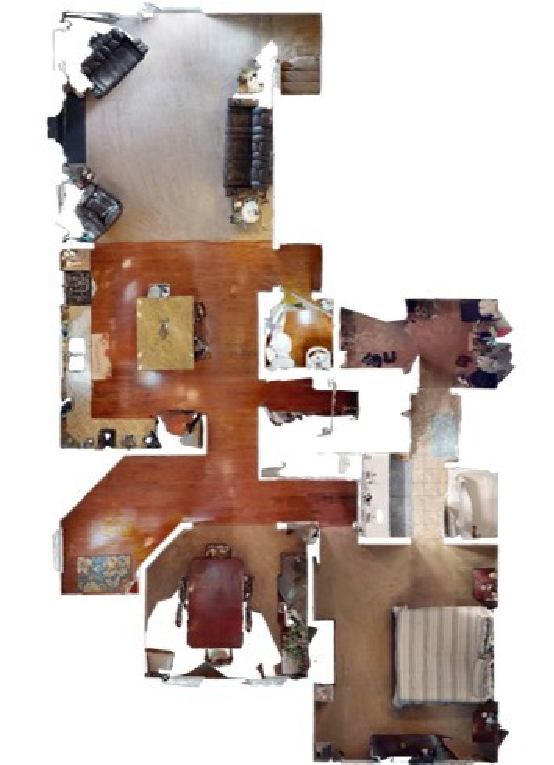}{00800-TEEsavR23oF}
\FIG{2.3}{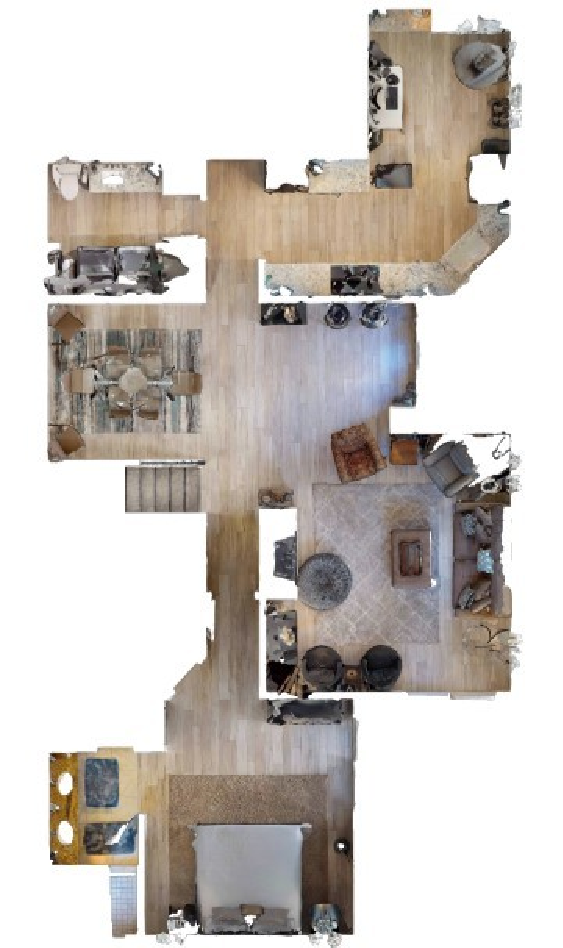}{00801-HaxA7YrQdEC}
\FIG{2.3}{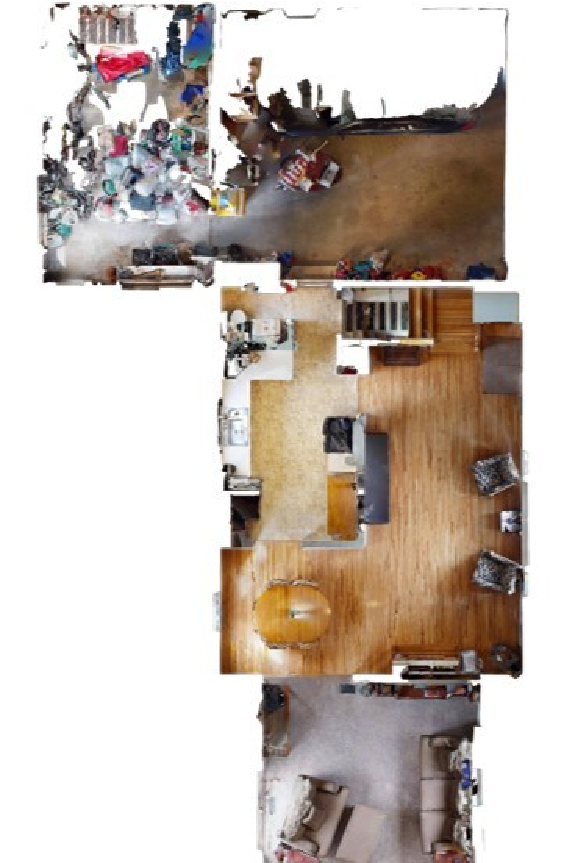}{00802-wcojb4TFT35}
\caption{%
Experimental environments.
}\label{fig:D}
\end{figure}
}

\newcommand{\figF}{
\begin{figure}[t]
\FIG{8}{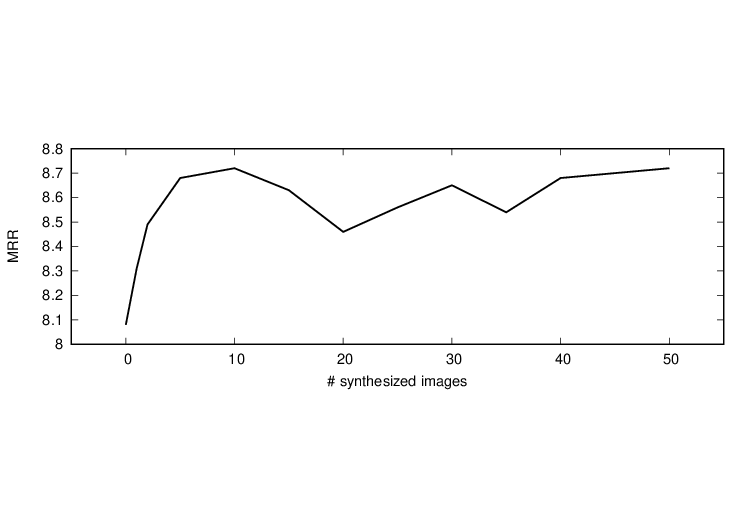}{}
\caption{%
Performance
versus 
the number of synthesized images.
}\label{fig:F}
\end{figure}
}

\newcommand{\tabA}{
\begin{table}[t]
\caption{%
Performance results.}
\arrayrulewidth=1.0pt 
 
\begin{tabular}{lccc} 
\hline
& \multicolumn{3}{c}{MRR} \\ \hline
& 00800 & 00801 & 00802 \\ \hline
semantic histogram  & 5.30 & 5.36 & 5.93 \\
NetVLAD  & 5.33 & 5.34 & 5.26 \\ 
Patch-NetVLAD & 7.28 & 6.59 & 8.21 \\
PNV-RRV  & 6.95 & 6.47 & 7.96 \\ 
PNV-RRV           VS & 7.58 & 7.06 & 7.59 \\ 
\hline
\end{tabular}
\vspace*{3mm}\\
{\scriptsize  00800: 00800-TEEsavR23oF,  00801: 00801-HaxA7YrQdEC, \\
 00802: 00802-wcojb4TFT35 \\
 RRV: reciprocal rank vector, VS: view synthesize 
}

\label{tab:A}
\end{table}
}

\newcommand{\tabB}{
\begin{table*}[t]
\caption{%
Timing results.
}\label{tab:B}
\mybox{3}{17}{
この表は、計算時間を載せる。
}
\end{table*}
}

\maketitle

\begin{abstract}
Cross-view self-localization is a challenging scenario of visual place recognition in which database images are provided from sparse viewpoints. Recently, an approach for synthesizing database images from unseen viewpoints using NeRF (Neural Radiance Fields) technology has emerged with impressive performance. However, synthesized images provided by these techniques are often of lower quality than the original images, and furthermore they significantly increase the storage cost of the database. In this study, we explore a new hybrid scene model that combines the advantages of view-invariant appearance features computed from raw images and view-dependent spatial-semantic features computed from synthesized images. These two types of features are then fused into scene graphs, and compressively learned and recognized by a graph neural network. The effectiveness of the proposed method was verified using a novel cross-view self-localization dataset with many unseen views generated using a photorealistic Habitat simulator.
\end{abstract}

\section{Introduction}

Cross-view self-localization is a challenging scenario of visual place recognition in which database images are provided from sparse viewpoints \cite{I9}. Most existing self-localization methods assume that database images similar to the query images have been encountered in the training stage. However, this assumption has frequently been proven false. For example, in autonomous driving scenarios, a vehicle often visits a place not only at different times of the day, weather, and seasons but also in opposite directions with vertically and horizontally shifted viewpoints. Under such severe viewpoint changes, most existing self-localization methods fail frequently \cite{I9}. Recently, NeRF (Neural Radiance Fields) and other view-synthesis techniques have introduced innovation to the cross-view self-localization problem \cite{I11}. For example, in \cite{I12}, NeRF-derived view synthesis effectively acts as a method for data augmentation in cross-view self-localization scenarios \cite{I13}. Our study was inspired by these achievements and aimed to explore this research direction.

\noeditage{
\figA
}

Despite their desirable properties, current view-synthesis approaches still encounter several challenges: 
(1) 
Storing a large number of synthesized images increases storage costs. To maintain real-time performance, a real-time system usually applies view synthesis to each database image rather than to the query image. Naively storing $N$ synthesized images per real database image requires $N$ times more storage than storing only real database images. One solution is to exploit the local similarities between real and synthesized images to compress them into a neural implicit representation (NIR), i.e., coordinate-to-feature map \cite{1238242}, which we intend to employ in the proposed approach. 
(2) 
Most existing view-synthesis techniques rely on the assumption that database images are available from multiple viewpoints spatially similar to the unseen viewpoint of the synthesized image to be generated. This is not the case in the cross-view self-localization scenarios. Typically, a robot visits each place in the workspace only once. One of the most relevant view-synthesis techniques is SynSin \cite{I14}
where only a single training image at a single viewpoint is required to generate an unseen view. However, in our preliminary experiments, this method often produces unnaturally synthesized images owing to GAN artifacts, particularly in cases where the unseen viewpoint is spatially distant from the real one (Fig. \ref{fig:C}).

In this study, we address a novel and challenging cross-view self-localization scenario, called ``single-shot" visual place classification, where only a single training image is provided for each place class (Fig. \ref{fig:A}). To solve this problem, we developed a highly accurate and efficient approach using graph neural networks. In particular, we chose a scene graph as our scene model because it is discriminative and compact and because it can generalize two major paradigms of scene model: global and local features \cite{I1}. The key idea was to introduce a new scene synthesis technique tailored for scene graphs. This allows the generation of synthesized scene graphs with similar scene parts (nodes) and different scene layouts (edges) at high speed and with high accuracy through lightweight graph manipulation. Furthermore, the scene graph of $N$ unseen viewpoints is compressed into the weight parameters of a graph convolutional neural network, rather than being stored directly in a spatially expensive manner. Finally, we present a new view-synthesis model that requires only a single-shot, place-specific training image, as in SynSin \cite{I14}. However, in contrast to SynSin, it is free of GAN artifacts. The proposed view-synthesis technique is based on the recently developed domain-invariant monocular depth estimation technique, MiDaS, in \cite{R8}, which does not require retraining, and is suitable for unsupervised robotics applications. The effectiveness of the proposed method is verified using a novel cross-view self-localization dataset with various views generated by a photorealistic Habitat simulator.

\noeditage{
\figC
}

The contributions of this study are summarized below. (1) We address a novel challenging ``single-shot" cross-view self-localization, where the query image's viewpoint may have undergone severe vertical and horizontal shifts as well as rotations in any direction. (2) The NeRF-based view synthesis was revisited in terms of computational efficiency, and a novel view-synthesis technique that was particularly tailored for scene graphs was developed by taking advantage of local and global features. (3) The proposed method clearly outperformed the baseline and ablation methods in evaluation experiments with various views generated by the photorealistic Habitat simulator.

\noeditage{
\figB
}

\section{Approach}

\subsection{Problem Formulation}\label{sec:problem}

Figure \ref{fig:B} shows the system overview. Our goal is to train a localization model for cross-view localization with multiple viewpoint changes from $N$ spatially sparse training images (or database images). Examples of such sparse viewpoints are shown in Fig. \ref{fig:B}. Suppose each of the $N$ training images is assigned a ground truth viewpoint position. Note that if training images from spatially dense viewpoints were available, a highly accurate structure-from-motion would be available, and then our problem would degenerate to the trivial task of model-based rendering from the reconstructed 3D model. In contrast, we consider a challenging scenario in which the training viewpoints are so sparse such a structure-from-motion on the training images is not applicable.

We formulated self-localization as a place classification task and constructed a dataset for a cross-view scenario \cite{A1}. A collection of $K=100$ place classes for each workspace defined and fixed throughout the experiment. Each place class is represented by a representative 2D point in the workspace, and all view images containing the representative point inside the visibility cone belong to that class. If a view image contained the  representative points of multiple places inside its visibility cone, it was regarded as belonging to the place with the closest physical distance from the viewpoint.

This setup is very challenging compared to existing studies on cross-view self-localization because not only opposite views but also views with all possible viewpoint orientations can belong to the same place class, which makes the resulting inner-class variance in the viewing direction significantly larger. Recognition becomes particularly difficult when the representative point of a place class is barely visible at the boundary of the visibility cone.

Images that did not belong to any place class were excluded from the dataset. It is known that humans have the ability to recognize places that we have never seen before (``Atmosphere-based self-localization"). However, such a challenging cross-view setup has not yet been explored, and remains an important issue for future studies.

\subsection{Scene Graph}\label{sec:scenegraph}

Two types of scene graph descriptions were used: global and part features. Global features describe the entire image using a single feature vector. This method is more discriminative, but may be vulnerable to viewpoint changes. For the part feature, we describe each part by one feature vector after semantically segmenting the image into scene parts. The semantic segmentation model in \cite{semseg} is employed. The graph node set consists of a whole image node and each part node. Its graph edge set consists of image-to-part edges connecting the whole image node and each part node, and part-to-part edges connecting the part pairs whose bounding boxes overlap.

Patch-NetVLAD \cite{R4} was used as the appearance descriptor. 
First, the scene is described using a bag of local features. Each local feature is then converted into a visual word with a score value. 

For feature-to-word conversion, we employed a dictionary of $k$ prototypes as the vocabulary. Recalling that only one training image is available per class, the set of training images of each class acts as one prototype. Then, each training image is described by a collection of PNV features.

Given the vocabulary, the process of describing a certain scene part is as follows: 
(1) calculate the degree of dissimilarity between the scene part and each prototype, 
(2) rank the prototypes in ascending order of dissimilarity, and 
(3) convert the result into a prototype-specific reciprocal rank vector (RRV) \cite{rrf}, 
which is then returned as a part descriptor.

For dissimilarity evaluation, the interset distance is calculated as the dissimilarity between the PNV set of a scene part and the PNV set of a prototype according to the naive Bayes nearest neighbor distance \cite{A3}. 

For the interset distance, the L2 norm between each scene part's PNV and each prototype's PNV was averaged over all scene parts' PNVs and then returned as the interset distance.

\subsection{Synthesizing Scene Graphs}

Understanding the 3D scene structure is the basis of NeRF techniques. For 3D scene reconstruction, we adopted a domain-invariant monocular depth estimation, MiDaS \cite{R8}, because it requires only a single-view training image and is a domain-invariant retraining-free model 
which we observe is suitable for our cross-view scenario. The 3D reconstruction procedure consisted of two steps. First, a depth image was generated using monocular depth estimation. The depth image was then mapped to a 3D point cloud using pinhole camera mapping. Unfortunately, this 3D point cloud was not of sufficient quality for the direct synthesis of synthesized images from other unseen viewpoints. Fortunately, it is still useful to features (e.g., Patch-NetVLAD) to other viewpoints, as discussed in the following subsections.

The simplest method to synthesize a scene graph is a two-step procedure of 
(1)
generating a synthesized image from a 3D point cloud 
and 
(2)
using it to extract the scene graph. However, this method presents challenges in terms of its quality and efficiency. As mentioned in the previous subsection, for the original real-view image, there is a one-to-one correspondence between each pixel and 3D point in the point cloud; however, this is not the case for the synthesized image. Consequently, the quality of a synthesized image has significant information loss compared to that of a real image. Furthermore, this approach requires costly synthesis operations $N$ times per image, making the time cost prohibitive.

Our approach aims to achieve scene-graph synthesis 
at each viewpoint
within the time budget provided by real-time localization. 
To do so, we decompose the image features to be synthesized into two categories: view-invariant features (e.g., appearance) and view-dependent features (e.g., spatial relationship), and then extract the different categories of features independently. For the former, 
we run the feature extraction process 
only once per training image using the original real-view images, whereas for the latter, we introduce a lightweight process that can be run for each synthesized view.

The RGB image (size:256$\times$256) is converted into semantic regions and depth images. The semantic regions are extracted using semantic segmentation (\ref{sec:scenegraph}). Depth images were generated using MiDaS \cite{R8}. 
The depth image was converted into a 3D point cloud using a pinhole camera model. This point cloud is then used to
calculate the synthesized image at another virtual viewpoint. 
The same pinhole camera model was used for this synthesis. The internal parameters of the pinhole camera model followed the intrinsic parameters from Habitat API Documentation \footnote{https://aihabitat.org/docs/habitat-api/view-transform-warp.html}.
Calibration was performed to match the depth direction (depth value) to the world coordinate system. 
Using the Z-buffer method from the projected image coordinates and depth information, 
an image was cropped according to the original image size (256$\times$256), 
and then, it is considered as a valid training image if the area of margin is less than 80\%.
A group of scene parts is generated based on the obtained virtual viewpoint semantic image and PNV keypoints, and then, a virtual viewpoint scene graph consisting of the entire image node and each part node is generated. The bounding box of the part node is calculated 
for each synthesized image. 
Note that this calculation can be very fast as it only needs to calculate the min-max pairs of coordinates along the $x$- and $y$- axes for each region.

\noeditage{

\figD

}

\section{Experiments}

We conducted experiments using the 3D photorealistic simulator Habitat-Sim \cite{E1}. The Habitat-Matterport3D Research Dataset (HM3D) was imported into Habitat-Sim. A bird's-eye view of the experimental workspaces are shown in Fig. \ref{fig:D}.

A challenging cross-view scenario was also considered. This dataset is challenging from several aspects. First, it consists of a size 100 class set represented by 100 representative points in the bird's-eye view coordinate system, implying that a viewpoint of any orientation that includes a place's representative point in the visibility cone, as in Fig. 4b, belongs to that place class (\ref{sec:problem}). Second, for each class, only a single training image was provided, which meant that many test views were very dissimilar to the training views. Third, the pairing of representative points with training images was chosen such that the 100 training images were similar to each other, which made it difficult to distinguish the training images from each other. Specifically, we used a simple single-cluster clustering technique to randomly sample 100,000 images into 100 similar subsets. At that time, the 3DOF robot pose space was divided by a three-dimensional grid (location resolution: 2m, azimuth resolution: 30deg) and the sampling was performed to ensure that two or more place classes did not belong to the same grid cell. For computational simplicity, we used randomly selected cluster centers rather than the costly k-means clustering. For the appearance similarity, the NBNN distance \cite{A3} using Patch-NetVLAD \cite{R4} was used as a dissimilarity measure. 

\noeditage{

\figE
}

The proposed method was compared with the baseline and ablation methods. 
Place classification performance was evaluated using mean reciprocal rank (MRR) \cite{E3}.
Recall that the proposed method uses node descriptors as hyperparameters. 
We also used semantic histogram \cite{E6} (in our implementation), which is a state-of-the-art semantic localization solution. A self-localization task using these features was cast into visual place classification using a dissimilarity function. For the dissimilarity functions, we used the L2 norm for NetVLAD, NBNN for Patch-NetVLAD, and inverse of the scoring function defined in the original study for the semantic histogram \cite{E6}.

Table \ref{tab:A} shows the performance results. As can be seen, the proposed method outperforms all the baseline and ablation methods considered. Among the ablation methods, the Patch-NetVLAD descriptor with a spatial edge description showed higher performance than its combination with other descriptors. This may be because the Patch-NetVLAD descriptor has already shown good performance, whereas the other descriptors have degraded discriminativity owing to the noise inherent in the synthesized images. Notably, the proposed method outperformed the case without view synthesis. From this result, we can conclude that the spatial edge description derived from synthesized images often contributes to performance improvement. The baseline method failed to achieve satisfactory performance in our challenging cross-view setup. NetVLAD is vulnerable to viewpoint changes as reported in other studies \cite{I2}. Patch-NetVLAD worked sufficiently well; however, in cases without view synthesis support, it did not make good use of the spatial information. A naive improvement to this method could be to introduce a reordering method, such as RANSAC post-verification, at an additional computational cost; however, such post-processing would boost the proposed method as well. The semantic histogram achieved impressive performance despite using only semantic features. However, its performance did not reach that of the proposed method, which combines all the semantic, spatial, and appearance features. From the above considerations, it can be concluded that the proposed method achieves robust self-localization under an adversarial cross-view setup with very sparse training images, and because it relies on a very simple scene graph descriptor, there is great potential for further performance improvements.

The relationship between the number of synthesized training images per place class and performance is shown in Figure \ref{fig:F}. Surprisingly, the maximum performance was already achieved when the number of part images was 10.

The overall processing time for one test image to generate a scene graph and classify places using a graph neural network was 0.328 [sec] (CPU: 11th Gen Intel(R) Core(TM) i7-11700K @ 3.60GHz, Python 3.6.9), which the ablation method without view synthesis costed 0.328 [sec].
The results indicate that the proposed method maintains the same computational speed as the baseline method, despite the fact that the former considers synthesized images at multiple viewpoints while the latter does not. One major reason is that the proposed method uses generation and learning of synthesized images for offline learning, and is designed to minimize the online computational load.

\noeditage{
\tabA
\figF
}

\section{Applications}

Nerf and other recent view synthesis technologies can be seen as variants of neural implicit representation (NIR). Assuming that a data instance comprises the pairs of a coordinate and its output features, INRs adopt a parameterized neural network as a mapping function from an input coordinate into its output features \cite{acm2013}. For more general NIR tasks, its use has recently been explored in the robotic mapping and localization community for other downstream tasks such as reconstruction, change detection, map compression, and map merging. The NIR technique proposed in this work can also be applied to these tasks as well. In particular, the proposed approach compactly and robustly describes new scenes by blending a set of prototype scene graphs. Compared to existing NIR techniques, the key novelty of the proposed method is that it targets topological maps rather than metric maps. Applying this approach to recent topological SLAMs such as graph neural SLAM and ego-centric topological maps is an urgent research topic.

\bibliographystyle{IEEEtran} 
\bibliography{reference}

\end{document}